\documentclass[11pt]{article} 

\usepackage[utf8]{inputenc} 

\usepackage{geometry} 
\usepackage{graphicx} 

\usepackage{booktabs} 
\usepackage{array} 
\usepackage{paralist} 
\usepackage{verbatim} 
\usepackage{subfig} 
\usepackage{dirtytalk}
\usepackage{wrapfig}
\usepackage{cite}

\usepackage{fancyhdr} 
\usepackage{sectsty}
\allsectionsfont{\sffamily\mdseries\upshape} 

\usepackage[nottoc,notlof,notlot]{tocbibind} 
\usepackage[titles,subfigure]{tocloft} 

\usepackage{changepage}
\usepackage[ruled,vlined,linesnumbered,noend]{algorithm2e}
\usepackage{amsmath,amsthm,amssymb}
\DeclareMathOperator{\sign}{sign}
\DeclareMathOperator*{\argmax}{argmax}

\title{\noindent\rule{17cm}{5.0pt} \\ 
	\bf Learning Control for Air Hockey Striking using Deep Reinforcement Learning \\
	\noindent\rule{17cm}{2.0pt} }
\author{\bf Ayal Taitler and Nahum Shimkin \\ Andrew and Erna Viterbi Faculty of Electrical Engineering \\ Technion Institute of Technology \\ Haifa,  Israel, 32000 \\ \{ataitler, shimkin\} @ technion.ac.il}
\date{} 
         
\begin{document}
\maketitle

\begin{abstract}
We consider the task of learning control policies for a robotic mechanism striking a puck in an air hockey game.
The control signal is a direct command to the robot's motors. We employ a model free deep reinforcement learning framework to learn the motoric skills of striking  the puck accurately in order to score. We propose certain improvements to the standard learning scheme which make the deep Q-learning algorithm feasible when it might otherwise fail. Our improvements include integrating prior knowledge into the learning scheme, and accounting for the changing distribution of samples in the experience replay buffer. Finally we present our simulation results for aimed striking which demonstrate the successful learning of this task, and the improvement in algorithm stability due to the proposed modifications. 
\end{abstract}

\section{\sc Introduction}
\label{introduction}
The problem of learning a skill, a mapping between states and actions to reach a goal in a continuous world, lies at the heart of every interaction of an autonomous system with its environment. In this paper, we consider the problem of a robot learning how to strike effectively the puck in a game of air hockey. Air hockey is a fast competitive game where two players play against each other on a low-friction table. Players are required to develop and perfect skills as blocking and striking in order to play and win. Classical approaches for striking the puck involve a multi stage process of planning and execution. First, planning a strategy based on the goal and skill, e.g., calculating the best point of collision to achieve the goal, then planning a path and trajectory and finally executing the low level motoric control \cite{namiki2013}.
Each part requires full knowledge of the mechanical and physical models, which might be complex. We propose doing the planning and the control simultaneously with learning, which offer an off model way to learn from the final result. The result will be given in a form of reward at the end of each trial, and will direct the learning to the correct policy. \\

Such problems include policy gradients \cite{williams1992simple} where a mapping between states and actions is learned with gradient ascent optimization on the accumulated reward, with or without keeping track of the value function. Another popular approach is Learning from Demonstration  (LfD) \cite{schaal1997learning, argall2009survey} sometimes refereed as imitation learning \cite{muelling2010learning} and apprenticeship learning \cite{abbeel2004apprenticeship}. In LfD a human expert (or a programmed agent) is recorded and the learning agent learns on the recorded data in a supervised fashion. Sometimes this process is used as an initialization for a second reinforcement learning stage for improvement. 
Paper \cite{bentivegna2002framework} used imitation learning to learn primitive behaviors for a humanoid robot in air hockey. \\

Exploration in such an environment is also an interesting issue. $\epsilon$-greedy exploration which is the most common one, is not highly efficient in such systems, since a dynamical system functions as a low pass filter \cite{kober2013reinforcement} and once in a while using a random action might have little affect on the output of the system. 
We combined several types of explorations including $\epsilon$-greedy and local exploration, along with prior knowledge based exploration that we proposed. \\

We propose an algorithm suitable for learning complex policies in dynamic physical environments.
The algorithm combined $\epsilon$-greedy exploration with a temporally correlated noise \cite{lillicrap2015continuous} for local exploration, which proved to be essential for effective learning. We further propose two novel contributions. We suggest a more relaxed approach to LfD which does not have the same limitations as standard LfD and can be learned from experience as regular RL. We also manage to overcome the instability of the learning due to the non-stationarity of the observed data, by expending the target update period.  \\ 

We compare our results with other deep reinforcement learning algorithms and achieve significant improvements, we are able to reach near optimal results, and keep them without suffering from a drop in the score function and the policies obtained.

\section{\sc Related Work}
\label{related}
Research on learning in autonomous systems was conducted in several directions. Our work has been influenced mainly from the recent work of deep Q-networks \cite{mnih2013playing, mnih2015human, silver2016mastering, van2015deep}, and the adaptation for continuous domains of deep deterministic policy gradients \cite{lillicrap2015continuous}. \\

Since the groundbreaking results shown by Deep Q-Learning for learning to play games on the Atari 2600 arcade environment, there has been extensive research on deep reinforcement learning. Deep Q-learning in particular seeks to approximate the Q-values \cite{watkins1992q}  using deep networks, such as deep convolutional neural networks. There has also been work on better target estimation \cite{van2015deep}, improving the learning by prioritizing the experience replay buffer to maximize learning \cite{schaul2015prioritized} and preforming better gradient updates with parallel batch approaches  \cite{mnih2016asynchronous, nair2015massively}. Some work on adaptation to the continuous control domain has been done also by \cite{lillicrap2015continuous}. Policy gradients methods were traditionally used \cite{williams1992simple, Kakade, peters2007reinforcement}, but struggled as the number of parameters increased. Adaptation to the deep neural network framework has also been done in recent years \cite{schulman2015trust, schulman2015high}. Several benchmarks such as \cite{duan2016benchmarking} have made comparisons between continuous control algorithms. In This paper we focus on the online DQN based approach, and extend it in the domain of continuous state optimal control for striking in air hockey.

\medskip

\section{\sc Deep Q-Networks} 
\label{dqn}
We consider a standard reinforcement learning setup consisting of an agent interacting with the environment in discrete time steps. At each step the agent receives an observation $s_t \in \mathbb{R}^n$ which represents the current physical state of the system, takes a action $a_t \in A$ which it applies to the environment, receives a scalar reward $r_t=r(s_t,a_t)$, and observes a new state $s_{t+1}$ which the environment transitions to. It is assumed that the next state is according to a stochastic transition model $P(s_{t+1}|s_t,a_t)$. The action set $A$ is assumed to be discrete. \\

The goal of the agent is to maximize the sum of rewards gained from interaction with the environment. Our problem is a finite horizon problem in which the game terminates if the agent reached some predefined time T. We define the future return at time $t$ as $R_t = \sum_{t'=t}^T{r_{t'}}$, where $T$ is the time at which the game terminates. The goal is to learn a policy which maximizes the expected return $\mathbb{E}\big[R_0\big]$ from the initial state. \\

The action-value function $Q^*(s,a)$ is used in many reinforcement learning algorithms. It describes the expected return after taking an action $a$ in state $s$ and thereafter following an optimal policy. The optimal state-action value function $Q^*$ obeys the equality known as the \textit{Bellman's equation}. 
\begin{equation}
Q^\ast\big(s_t,a_t\big) = \mathbb{E}_{s_{t+1}}\Big[r_t + \max_{a'}Q^*\big(s_{t+1},a') \Big| s_t,a_t \Big]
\end{equation}
For learning purposes it is common to approximate the value of $Q^\ast\big(s,a\big)$ by using a function approximator, such as a neural network. We refer to the neural network function approximator with weights $\theta$ as a Q-network. A neural network representing the Q-function can be trained by considering the loss function:
\begin{equation} \label{eq:loss}
L\big(\theta\big) = \mathbb{E}_{s_t,a_t,r_t,s_{t+1} \sim D}\Big[\Big(y(\theta) - Q\big(s_t,a_t ; \theta\big) \Big)^2 \Big]
\end{equation} 
where
\begin{equation}
\begin{aligned}
&y(\theta) = r\big(s_t,a_t\big) && s_{t+1} \text { terminal} \\
&y(\theta) = r\big(s_t,a_t\big) + \displaystyle \max_{a} Q\big(s_{t+1},a ; \theta \big) && s_{t+1} \text { not terminal}
\end{aligned}
\end{equation}
During training time each transition of state, action, reward and next state $<s_t,a_t,r_t,s_{t+1}>$ is stored in an \textit{experience replay buffer} $D$ from which samples are drawn uniformly in order to reduce time correlations to train the network. $y(\theta)$ is called the target and typically also a function of $\theta$. The $\text{max}\{\cdot\}$ operator in the target makes it hard to calculation derivatives in respect to the weights, so the target is kept constant and the derivatives are calculated only according to $Q(s_t,a_t;\theta)$. This loss function has the tendency to oscillate and diverge. In order to keep the target stationary and prevent oscillations, the DQN algorithm make use of another network, called a target network with parameters $\hat \theta^-$. The target network is the same as the on-line network except that its parameters are copied every $C$ updates from the on-line network, so that $\hat \theta^-$ are kept fixed during all other updates. The training of the network in this case is according to the following sequence of loss functions
\begin{equation}
L_i\big(\theta_i\big) = \mathbb{E}_{s_t,a_t,r_t,s_{t+1} \sim D}\Big[\Big(y_i(\hat \theta^-_i) - Q\big(s_t,a_t ; \theta_i\big) \Big)^2 \Big]
\end{equation}
The target used by DQN is then 
\begin{equation}
y_i(\hat \theta^-_i) = r\big(s_t,a_t\big) + \displaystyle \max_{a} Q\big(s_{t+1},a ; \hat \theta^-_i \big)
\end{equation}
and the on-line network weights can be trained by stochastic gradient descent (SGD) and back-propagation
\begin{equation}
\theta_{i+1} = \theta_{i} + \alpha \nabla_{\theta_i} L_i(\theta_i)
\end{equation}
where $\alpha$ is the learning rate. An improvement on that has been proposed in the double DQN algorithm, the decoupling of the estimation of the next value and the selection of the action, and decrease the problem of value overestimation, the following target has been used
\begin{equation}
\begin{aligned}
&y_i(\hat\theta^-_i) = r\big(s_t,a_t\big) +  Q\Big(s_{t+1} ,  a_{t+1} ; \hat \theta_i^- \Big) \\
&a_{t+1} = \argmax_a  Q\big(s_{t+1},a ; \theta_i \big)
\end{aligned}
\end{equation}
In our work unless specified otherwise all learning updates have been done according to the double DQN learning rule. \\

To explore the environment the systems typically explore via the $\epsilon$-greedy heuristic. Given a state, a deep Q-network (DQN) predicts a value for each action. The agent chooses the action with the highest value with probability $1-\epsilon$ and a random action with probability $\epsilon$.

\medskip
 
\section{\sc Striking in Air Hockey}

We next introduce the striking problem and our learning approach.
\subsection{\sc The Striking Problem} \label{description}
The striking problem deals in general with interception of a moving puck and striking it in a controlled manner. We specialize here to the case where the puck is stationary. We wish to learn the control policy for striking the puck such that after the impact, the puck trajectory will have some desired properties. 
We focus on learning to strike the puck directly to the opponent's goal. We also considered some other different modes of striking the puck, Such as hitting the wall first. These are not presented here, but the same learning scheme fits them as well. We refer to these modes as skills, which a high level agent can choose from in full air hockey game.
The learning goal is to be able to learn these skills with the following desired properties
\begin{itemize}
\item the puck's velocity should be maximal after the impact with the agent.
\item the puck's end position at the opponent's side should be the center of the goal.
\item the puck's direction should be according to the selected skill.
\end{itemize}
The agent is a planar robot with 2 degrees of freedom, X and Y (gantry like robot). We used a second order kinematics for the agent and puck.
The state vector of the problem is $s_t \in \mathbb{R}^8$, which includes all the position and velocities of the agent and the puck in both axes, i.e., $s_t = \big[m_x,\ m_{Vx},\ m_y,\ m_{Vy},\ p_x,\ p_{Vx},\ p_y,\ p_{Vy} \big]^T$. Here $m_{*}$ stands for the agent's state variables and $p_*$ stands for the puck's state variables. The actions are $a_t \in \mathbb{R}^2$, and include the accelerations in both axes for the agent. \\

The striking problem can be described as the following discrete time optimal planning problem:
\begin{equation}
\begin{aligned}
& \underset{a_k}{\text{minimize}}
& & \phi(s_T,\ T) \\ 
& \text{subject to}
& & s_{k+1} = f(s_k,a_k) \\
&&& s^{(i)}_k \in \big[S^{(i)}_{min},\ S^{(i)}_{max} \big],\ \quad i=1,\ldots,8 \\
&&& a^{(j)}_k \in \big[A^{(j)}_{min},\ A^{(j)}_{max} \big], \quad j=1,2 \\
&&& s_0 = s(0) \\
\end{aligned}
\end{equation} 
Here the objective function $\phi(s_T,\ T)$ represents the value of the final state $s_T$ (in terms of velocity and accuracy), and the final time $T$ which we desire to be small. The function $f(\cdot)$ is the physical model dynamics. $S^{(i)}_{min},\ S^{(i)}_{max}$ and $A^{(j)}_{min},\ A^{(j)}_{max}$ are the constraints on the state (table boundaries and velocities) and action spaces (accelerations/torques) respectively. $s_0$ is the initial state.
We assume that $f(\cdot)$, the collision models and the table state constraints are hidden from the learning algorithm, The best known collision model is non-linear and hard to work with \cite{partridge2000control}. Solving analytically such a problem when these function are known is a challenging problem, when they are unknown it is practically impossible with analytic tools. In the simulations specific models were specified as explained in Section \ref{experiments}. \\

In order to fit the problem as stated in \ref{description} to the DQN learning scheme, where the outputs are discrete Q values associated with discrete actions, we discretized the action space by sampling a $2D$ grid with $n$ actions in each dimension (each dimension represents an axis in joint frame). Thus, we have $n^2$ actions. We make sure to include the marginal and the zero action, so our class of policies we search in will include the Bang-Zero-Bang profile which is associated with time optimal problems. Each action is associated with an output of the neural network, where each output represents the Q-values of each action under the input state supplied to the network, e.g., if state $s$ is supplied to the network, output $i$ is the Q-value of $Q(s,a_i;\theta)$. Thus, for every given state we have $n^2$ Q-values from the network, associated with the $n^2$ actions.

\subsection{\sc Reward Definition}
\label{reward_shaping}
The learning is episodic and in this problem the agent receives success indication only upon reaching a terminal state and finishing an episode. The terminal states are states in which the mallet collide with one of the walls (table boundaries violation), and the states in which the mallet strikes the puck (the agent does not perform any actions beyond this point). Any other state including the states in which an episode terminates due to reaching the maximal allowed steps, are not defined as terminal states.
At the terminal state of each episode the agent receive the following reward
\begin{equation}
R_{terminal} = r_c + r_v + r_d
\end{equation}
$R_{terminal}$ is consists of three components. The first is $r_c$, which is a fixed reward indicating a puck striking. The second component is a reward which encourages the agent to strike the puck with maximum velocity, and given by
\begin{equation}
r_v = \sign{(V)} \cdot V^2
\end{equation}
where $V$ is the projection of the velocity on the direction of a desired location $x_g$ on the goal line of the opponent. The last component is a reward for striking accuracy, which indicates how close the puck reached  $x_g$.
\begin{equation}
r_d = 
\begin{cases}
c & |x-x_g| \leq w \\
c \cdot e^{-d \cdot (|x-x_g|-w)} & |x-x_g| > w
\end{cases}
\end{equation}
where $x$ is the actual point the puck reaches on the opponent's side on the goal line, $c$ is a scaling factor for the reward, $w$ is the width of the window around the target point which receives the highest reward and $d$ is a decay rate around the desired target location. Naturally, if the episode terminates without striking the puck $R_{terminal}$ is zero.
In order to encourage the agent to reach a terminal state in minimum time, the agent receives a negative small reward $-r_{time}$ for each time step of the simulation until termination. The accumulative reward for the entire episode then is $R_{total} = R_{terminal} - n \cdot r_{time}$, where $n$ is the number of time steps for that episode.

\subsection{\sc Exploration}
The problem of Exploration is a major one, especially in the continuous domain. We address the issue from two angles, completely random exploration and local exploration.
\subsubsection{\sc Completely Random Exploration}
We use $\epsilon$-greedy exploration (see Section \ref{dqn}) in order to allow experimenting with arbitrary actions. In physical systems with inertia it is not efficient since the system acts as a low pass filter, but it does give the agent some sampling of actions it would not try under normal conditions.
\subsubsection{\sc Local Exploration}
The main type of exploration is what we refer to as \textit{local} exploration. Similarly to what was done in \cite{lillicrap2015continuous}, we added a noise sampled from a noise process $\mathcal{N}$ to our currently learned policy.Since the agent can apply only actions from a discrete set of actions $\mathcal{A}$, we projected the outcome on the agent's action set:
\begin{equation}
a_t = \mathcal{P}_{\mathcal{A}} \{ \argmax_a Q\big(s_t,a ; \theta \big) + \mathcal{N}_t  \}
\end{equation}
We used for $\mathcal{N}$ an Ornstein\textendash Uhlenbeck process \cite{uhlenbeck1930theory} to generate temporally correlated exploration noise for exploring efficiently. The noise parameters should be chosen in such a way that after the projection the exploration will be effective. Small noise might not change the action after the projection, but large noise might result in straying too far from the greedy policy. Thus, the parameters of the noise should be in proportion to the actions range and the aggregation.

\subsection{\sc Prior Knowledge from Experience}
In a complex environment, learning from scratch has been shown to be a hard task. Searching in a continuous high dimensional spaces with local exploration might prove futile. In many learning problems prior knowledge and understanding are present and can be used to improve the learning performance. A common way of inserting priors into the learning process uses LfD. 
For that purpose, multiple samples of expert performance should be collected, which is not always feasible or applicable. \\

In many cases the prior knowledge can be translated to some reasonable actions, although usually not an optimal policy.
Examples for that can be seen in almost every planning problem. In games, the rules give us some guidance to what to do, e.g., in soccer, \say{Kick the ball to the goal}, so for an agent to spend time on learning the fact that it has to kick the ball is a waste. In skydiving, the skydivers are told to move their hands to the sides in order to rotate, they are not required to search every possible pose to learn how to rotate. Furthermore, the basic rotating procedure taught to new skydivers is not the correct way to do it, it is taught as a basic technique, an initialization for them to modify and find the correct way. \\

We propose showing the agent a translation of the prior knowledge as a teacher policy. In some episodes instead of letting the agent to act according to the greedy policy, it does what the teacher policy suggests. The samples collected in those episodes are stored in the experience replay buffer as any other samples, allowing the learning algorithm to call upon that experience from the replay buffer and learn from it in within the standard framework. \\

For the problem of air hockey, we used a policy encapsulating some crude knowledge we have of the problem. We just instruct the agent to move in the direction of the puck, regardless of the task at hand (aiming to the right\textbackslash left\textbackslash middle), since this knowledge was simple, and robust enough.
The guidance policy we constructed has the following form:
\begin{equation}
\begin{aligned}
&V_{next} = \frac{P_{puck} - P_{agent}}{\| P_{puck} - P_{agent} \|} \cdot MaxVelocity \\
&a = \frac{\frac{V_{next} - V_{agent}}{\Delta t}}{\|\frac{ V_{next} - V_{agent}}{\Delta t}\|} \cdot MaxForce
\end{aligned}
\end{equation}
where $P_{object}$ is the $x,y$ position vector of the object, and $MaxVelocity,\ MaForce$ are physical constraints of the mechanics. The agent acts by the projection of the policy on its action space $\mathcal{P}_{\mathcal{A}} \{\cdot\}$ This policy will result with an impact between the agent and puck, but by no account will be considered as a \say{good} strike since there is no reason the puck will move in the goal's direction (except in the special case when the puck lays on the line between the agent and the goal). 
The guidance policy is shown (the agent acts by it) and stored in the replay buffer with probability $\epsilon_p$. \\

\subsection{\sc Non-Uniform Target Network Update Periods}
The deep reinforcement learning problem is different from the supervised learning problem in a fundamental way, as the data which the network uses during the learning changes over time. At the beginning, the experience replay buffer is empty, the agent starts to act and fills the buffer, when the buffer reaches its maximal capacity new transitions overwrite the older ones. It is obvious that the data is changing over time, first changing in size and then changing in distribution. As the agent learns and gets better, the data in the buffer reflects that and the samples are more and more of good states which maximize the reward. \\

Recall that the value the neural network tries to minimize is the loss function stated in (\ref{eq:loss}). In order to stabilize the oscillations a target network with fixed weights over constant update periods were introduced. That led to the stationarity of the target value. The choosing of update period length became of the parameters that had to be set. Small update period result with instability since the target network changes too fast and oscillates, large update periods may be too stationary and the bootstrap process might not work properly. Thus, a period that is somewhere in the middle is chosen so the updates are stable.\\

In may domains such as in the air hockey and also in some of Atari games, DQN still suffers from a fall in the score. We argue that this fall is not only due to value overestimation (it happens for Double DQN updates as well), but also for issues with the target value. Choosing a middle value for the update period may result in slow learning in the beginning and fall in the score later in the learning due to oscillations. \\
  
In many domains such as in the air hockey and also in some of Atari games, DQN still suffers from a drop in the score as the learning process progresses (see, e.g., Fig. \ref{fig:DDQN}). We argue that this drop is not only due to value overestimation (it happens for Double DQN updates as well), but also for issues with the target value. Choosing a middle value for the update period may result in slow learning in the beginning and a drop in the score later in the learning due to oscillations. \\

We show that by adjusting the update period over time, we manage to stabilize the learning and prevent completely the drop in the score. We start with a small update period since the replay buffer $D$ is empty and we want to learn quickly, we then keep expanding the period as the buffer gets larger, and we need more sampling to cover it. As the agent gets better and the distribution stabilizes, we also expand the update period in order to filter oscillations and keep the agent in the vicinity of the good learned policy. The expansion of the update period is done at each target weights update according to 
\begin{equation}
C = C\cdot C_r, \qquad C_r \geq 1
\end{equation}
where $C_r$ is the expansion rate. When $C_r=1$ the updates are uniform as in the standard DQN. \\

At the beginning every sample contains new information that should affect the learning. As the learning progresses and the optimal policy hasn't been obtained yet, the samples in the replay buffer are diverse allowing the agent to learn from good samples and bad samples as well. At later stages when the agent has already learned a good policy, and the distribution of samples in the replay buffer resembles that policy. The network at the point if learning continuous, might suffer from what is known as the catastrophic forgetting \cite{french1999catastrophic} of neural networks. Freezing the target network before that stage, stabilize the learning and allows the network fine tune its performances, even though the distribution in the replay buffer is undiverse. The target network contains then the knowledge gained in the past from bad examples. At that stage of the learning the update period should be large for that purpose. This is achieved by gradually increasing the update period from an initial small period at the beginning during the learning.

\subsection{\sc Guided-DQN}
Putting the above-discussed features together produces the guided-DQN algorithm we used in the air hockey problem. The algorithm is given in algorithm \ref{alg:gDQN}. \\
\begin{algorithm}[h!] 
\caption{Guided Deep Q-Network \label{IR}}\label{alg:gDQN}
\SetKwInOut{Input}{input}\SetKwInOut{Output}{output}
\DontPrintSemicolon
\SetAlgoNoLine
\BlankLine
\Input{Guidance policy $\pi(s)$, Expansion rate $C_r$}
Initialize replay memory $D$ \\
Initialize states-actions value function $Q$ with random weights $\theta$ \\
Initialize target states-actions value function $\hat{Q}$ with weights $\hat \theta^-=\theta$ \\
\For {episode=1,M}{
	Observe initial state from environment $s_0$ \\
	Initialize random process $n$ for action exploration\\
	with probability $\epsilon_p$ decide if this episode is guided or not\\
	\While {t$<$N and $s_t$ is not terminal}{
		\If {guided episode} {
			Select $a_t = \displaystyle \mathcal{P}_\mathcal{A}\{ \argmax_a\ \pi\big(s_t\big) \}$
		}
		\Else {
			With probability $\epsilon$ select random action $a_t$ otherwise select $a_t = \mathcal{P}_\mathcal{A} \{\displaystyle \argmax_a \ Q(s_t,a;\theta) + n_t \}$
		}
		Execute action $a_t$ in environment and observe reward $r_t$, next state $x_{t+1}$ and if terminal $d_{t+1}$\\
		Set new state $s_{t+1} = s_t,x_{t+1}$ \\
		Store transition $<s_t,a_t,r_t,s_{t+1}>$ in $D$ \\
		Sample random mini-batch of transition $<s_j,a_j,r_j,s_{j+1}>$ from D \\
		set $y_t=r_t +  Q\big(s_{j+1} ,\displaystyle \argmax_a \ Q\big(s_{j+1},a ; \theta \big) ; \hat \theta^- \big)$ \\
		Perform gradient descent step on $\big( y_j-Q(s_j, a_j ; \theta )  \big)^2$ with respect to the network parameters $\theta$\\
		Every $C$ step reset $\hat{Q} = Q$, and set $C = C \cdot C_r$
	}
}
\KwRet{$Q$}
\end{algorithm}

As an input the algorithm gets the guidance policy, which encapsulated the prior knowledge we have on the problem, and the expansion rate $C_r$. At each
episode, with probability $\epsilon_p$ the entire episode will be executed with the guidance policy $\pi(s)$, or with probability $1-\epsilon_p$ according to the greedy policy, with the addition of time correlated exploration noise. In either case, a guided episode or a greedy episode, at each step the algorithm stores the transitions in the replay buffer, and preforms a learning step on the Q network. Samples from the replay buffer are selected randomly with uniform probability. The projection operator $\mathcal{P}_A$ projects the continuous actions onto the agent's discrete set, by choosing the action with the lowest euclidean distance. Every $C$ updates the target Q network is updated with the weights of the on-line Q network, and $C$ is expanded with a factor of $C_r$ so the next time the target network gets updated, it will be after a longer period than the previous update. \\

The learning rule is a Double DQN learning rule. Note that if the algorithm is not provided with a guidance policy (equivalent to setting $\epsilon_p$ to zero), $C_r = 1$, and the temporal correlated process is $\mathcal{N} \equiv 0$, the GDQN algorithm reduces to the standard Double DQN algorithm.

\medskip

\section{\sc Experiments}
\label{experiments}

\begin{figure}[!h]
\centering
\framebox{
\includegraphics[scale=0.4]{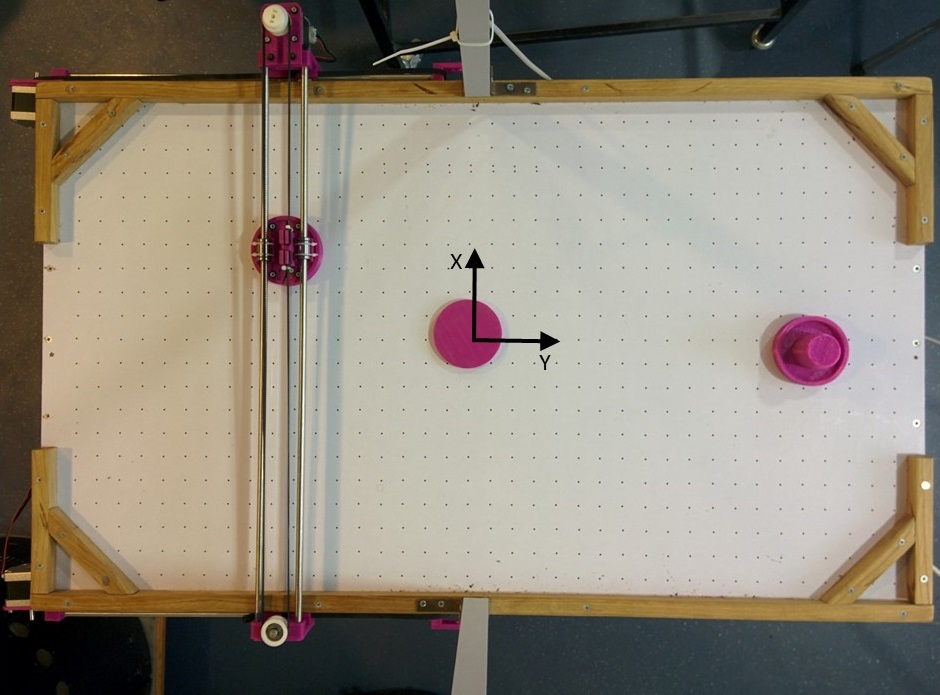}
}
\caption{The air hockey robotic system.}
\label{fig:robot}
\end{figure}

The simulation was fashioned after the robotic system in Fig. \ref{fig:robot}. 

In the robotic system the algorithm would learn on the real unknown physical models, but for the purpose of simulation we used simulation models for the agent dynamics and collision models. The simulation models are hidden from the learning algorithm and exist solely for the purpose of simulating the system for learning. For the agent dynamics we used a discrete time second order dynamics
\begin{equation}
\label{eq:dynamics}
\begin{split}
\begin{bmatrix}
X_m \\ V_{x,m} \\ Y_{m}\\ V_{y,m} 
\end{bmatrix}_{k+1} &= 
\begin{bmatrix}
1 & T & 0 & 0 \\
0 & 1 & 0 & 0 \\
0 & 0 & 1 & T \\
0 & 0 & 0 & 1 
\end{bmatrix}
\begin{bmatrix}
X_m \\ V_{x,m} \\ Y_{m}\\ V_{y,m} 
\end{bmatrix}_{k} + 
\begin{bmatrix}
0 & 0 \\
T & 0 \\
0 & 0 \\
0 & T
\end{bmatrix}
\begin{bmatrix}
a_x \\ a_y
\end{bmatrix}_{k} \\
\end{split}
\end{equation}
under the following constraints
\begin{equation*}
\begin{split}
|a_{x,y}| &< \text {Maximum\ force} \\
|V{\{x,y\},m} | &< \text{Maximum velocity} \\
|{X_m,Y_m}| &< \text{Table boundaries}
\end{split}
\end{equation*}
These constraints represent the physical constraints present in the mechanical system, where the velocity has a maximum value, the torques are bounded and we are not allowing the mallet to move outside of the table boundaries. \\

We used in the simulations an ideal impact model between the mallet and puck in the sense that we neglected the friction between the bodies during the impact and we assume the impact is instantaneous with energy loss according to a restitution coefficient $e$. The forces, accelerations, velocities and space (the field's boundaries) are constrained to reflect the physical constraints in the robotic system. \\

The list of parameters (learning rate, probabilities, etc.) used throughout the simulations is given in table \ref{tbl:parameters}.
\begin{table}[!h]
\caption{List of hyper-parameters used by the learning algorithm}
\label{tbl:parameters}
\begin{center}
\begin{tabular}{|c|c|c|}
\hline
Parameter & Value & Description \\  \hline
$\alpha$ & 0.00025 & learning rate \\
$\epsilon$ & 0.1 & $\epsilon$-greedy exploration probability\\
$\epsilon_p$ & 0.1 & policy demonstration probability \\
D & $2 \cdot 10^5$ & replay buffer size \\
N & 300 & episode's max steps \\
A & $ 5 \times 5 $ & action space size\\
Mini-Batch & 64 & mini batch size sampled from buffer \\
$C$ & expanding & update rate of Q target network \\
$C_r$ & 1.2 & expansion rate \\
\hline
\end{tabular}
\end{center}
\end{table}

The learning environment is a custom built simulation based on OpenAI gym \cite{brockman2016openai}. The simulation is modeled after an air hockey robotic system with two motors and track, one for each axis. The simulation includes visually the table, the mallet and the puck. \\

We simulated each attempt to strike the puck as an independent episode comprised of discrete time steps. At the beginning of each episode the puck is placed at a random position on the table at the agent's half court with zero velocity and the agent starts from a home position (a fixed position near the middle of the goal) with zero velocity. Each episode terminates upon reaching a terminal state or upon passing the maximum number of steps defined for an episode. The maximum steps number is 150 steps and the terminal states are the states where the agent collides with the puck (\say{good} states) or with one of the walls (\say{bad} states). The environment returns a reward as described in Section \ref{reward_shaping}. No reward is given upon hitting a wall beyond the timely reward. \\

The dynamic model of the puck and agent is a second order model as described in Section \ref{description}. $T$ is the sampling time of the system and was set to $0.05\ [sec]$ in the simulation. The puck's rotation was neglected, thus the collision models (puck-agent, puck-wall) are ideal with inbound and outbound angles the same. Energy loss in the collisions was modeled with restitution coefficient of $e=0.99$.  \\

The controller is a non-linear neural controller, a fully connected  Multi-Layer Perceptron with 4 layers (3 hidden layers and an output layer), the first two hidden layers are of 100 units each, the third hidden layer is of 40 units and the output layer is of 25 units. All activations are the linear rectifier $f(x)=\max(0,x)$. The controller is a map between states $s_t$ (the inputs to the controller) and discretized Q-values. We choose 5 actions in each axis, yielding $25$ output  actions\textbackslash Q-values (see Section \ref{description}). We used the RMSProp algorithm with mini-batches of size 64. \\

In all the simulation experiments we measured the score for random initial positions of the puck, it will always be shown in graph with the caption random. In addition we measured the performances for additional 3 fix representing states of the puck, fixed positions in the left side, the middle and the right side of the table. In addition we estimated the average value of all the states and present it as well.  The graphs matching these measures will be shown with appropriate captions. We present in this paper the results for the \say{direct hit}. \\

\subsection{\sc Results}
\begin{figure*}[!b]
      \centering
      \framebox{
\includegraphics[scale=0.22]{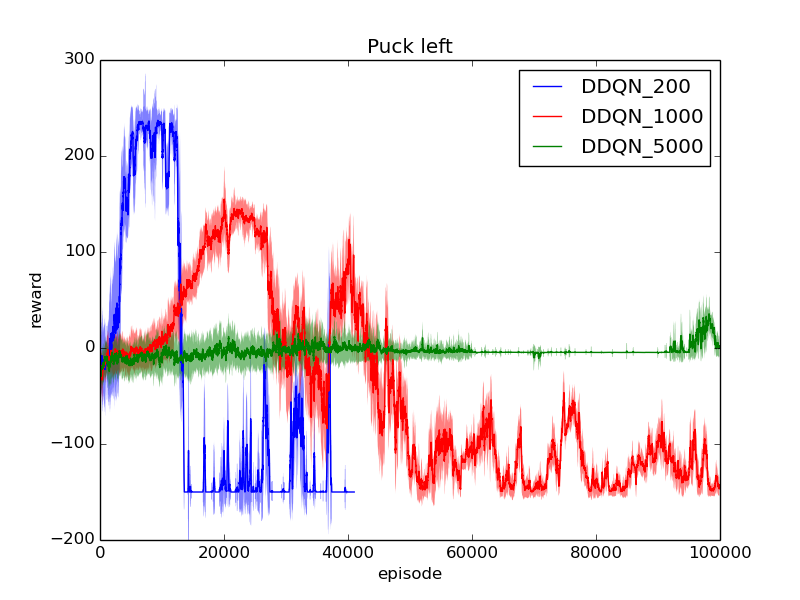}
}
\framebox{
\includegraphics[scale=0.22]{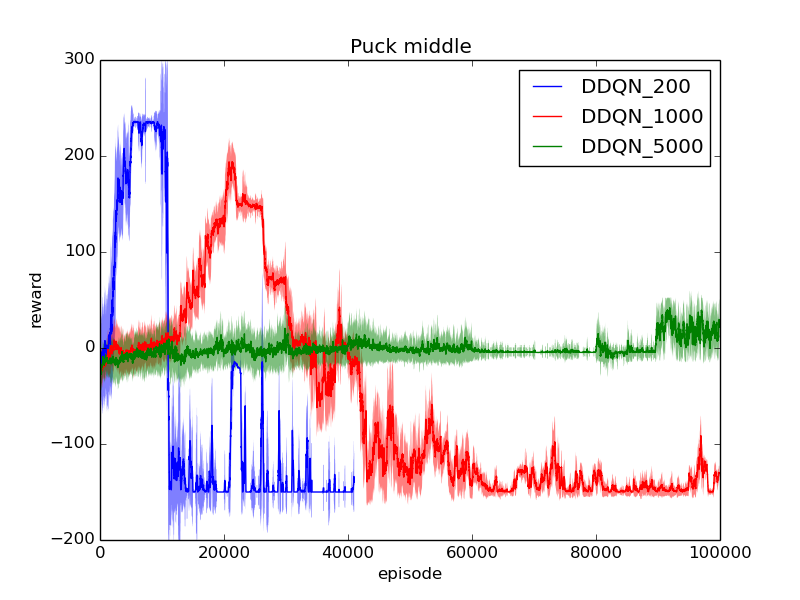}
}
\framebox{
\includegraphics[scale=0.22]{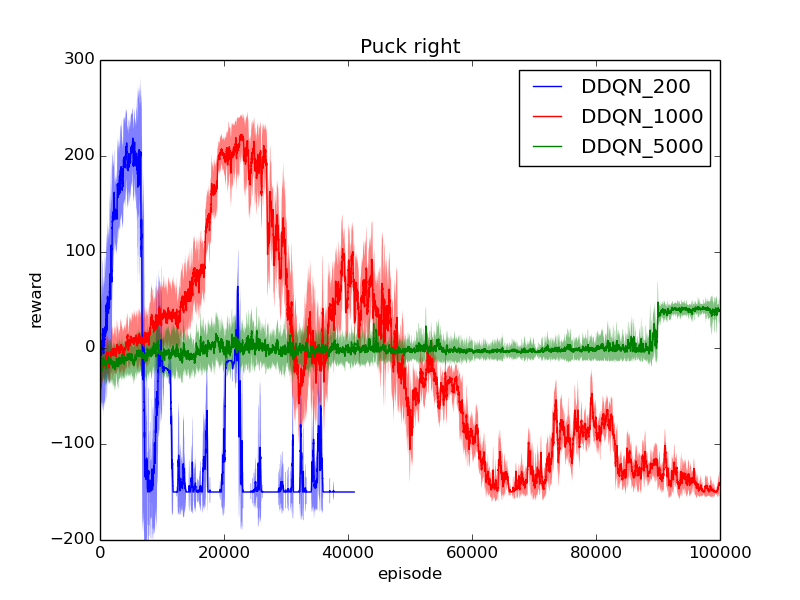}
}
\\
\framebox{
\includegraphics[scale=0.22]{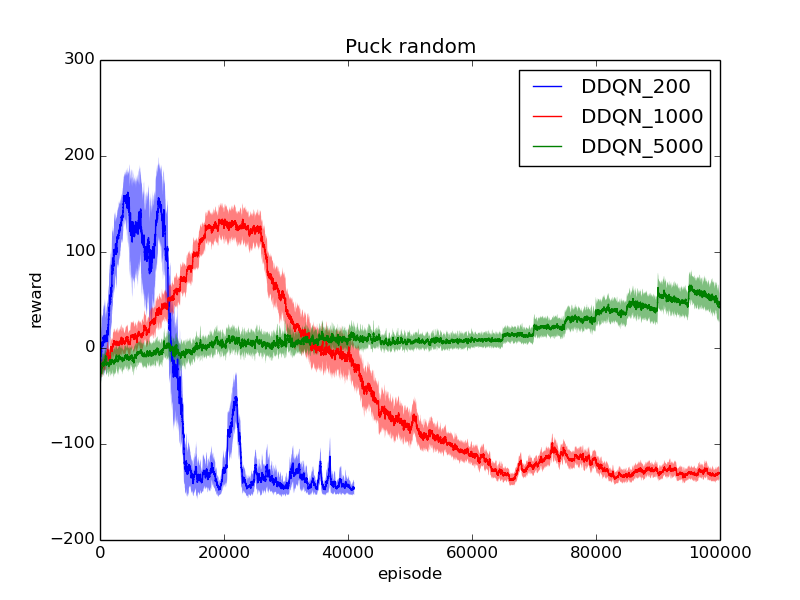}
}
\framebox{
\includegraphics[scale=0.22]{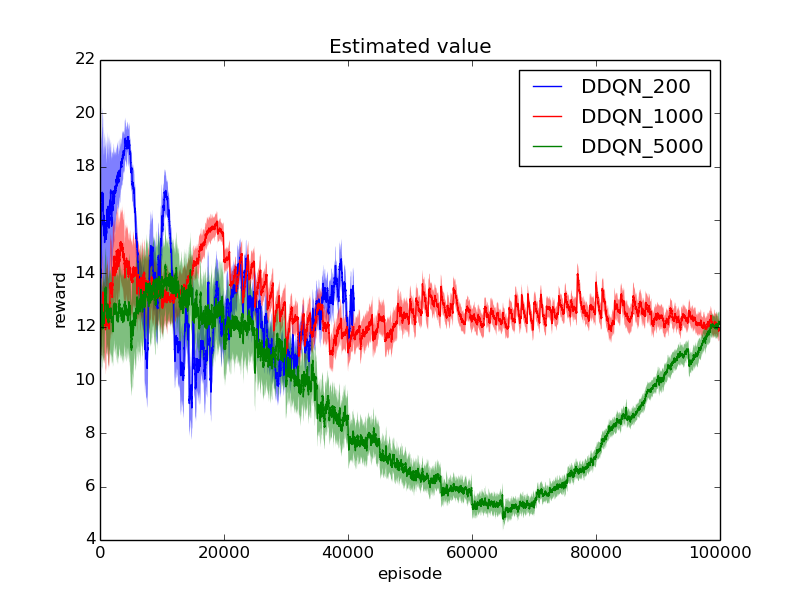}
}
      \caption{Double DQN results for the air hockey striking problem.}
      \label{fig:DDQN}
\end{figure*}

First we show the performance of the standard Double DQN in Fig. \ref{fig:DDQN} for different target network update periods. We choose a fast period a intermediate period and a slow period calculated such that each state in the buffer will be visited 8 times on average before being thrown away from the buffer. \\

It can be seen that the DDQN with fast updates ($DDQN_{200}$) rises the fastest but also drops quickly, the same behavior can be observed for the intermediate updates ($DDQN_{1000}$) but the rise is slower and  the drop happens less sharply. The score value the network drops to, $-150$, is exactly the value of the time penalty for a complete episode, i.e., the agent doesn't reach a terminal state. When investigating the policies obtained it can be seen that the agent's action oscillated between two opposite actions which affectively cause it to stand still.
For the slow updates ($DDQN_{5000}$) the case is different, the network seems mostly indifferent to the updates, and at the end it manages to rise a little.
The average value for all three runs oscillates and in general suffers from severe underestimation. \\

\begin{figure*}[!t]
      \centering
      \framebox{
\includegraphics[scale=0.22]{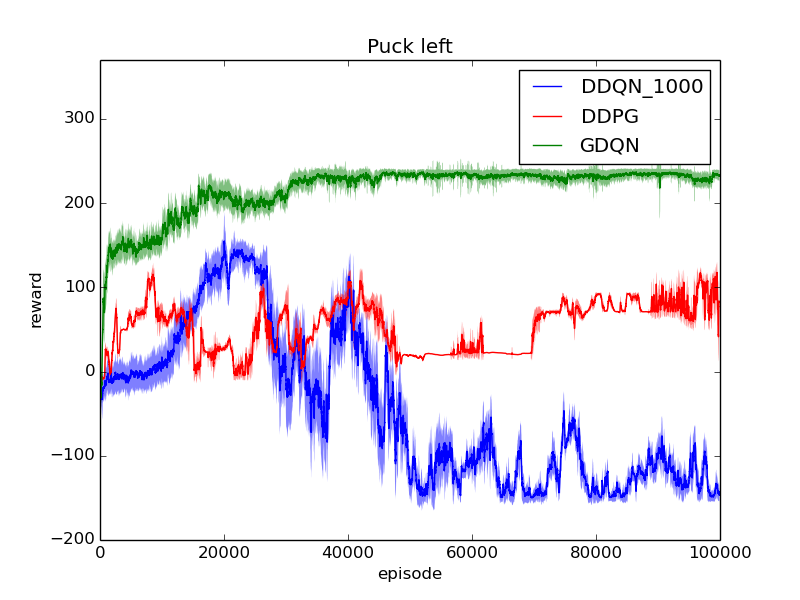}
}
\framebox{
\includegraphics[scale=0.22]{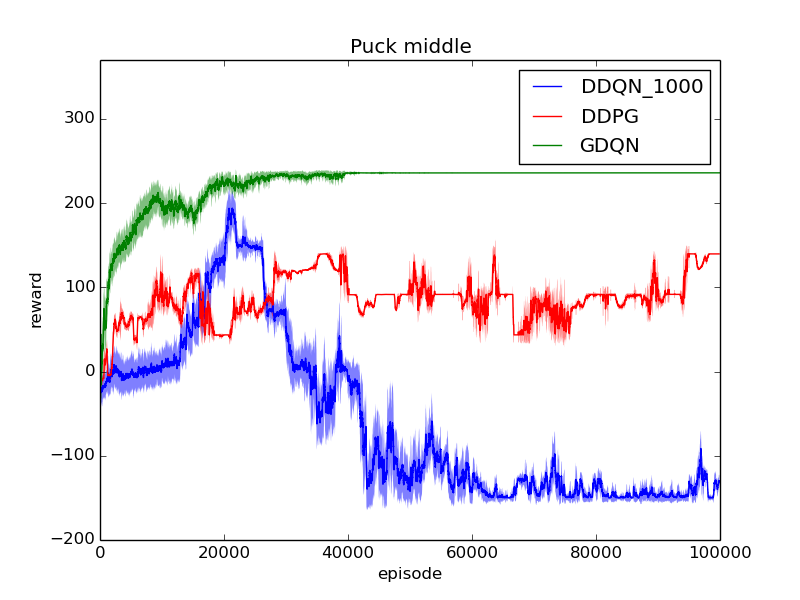}
}
\framebox{
\includegraphics[scale=0.22]{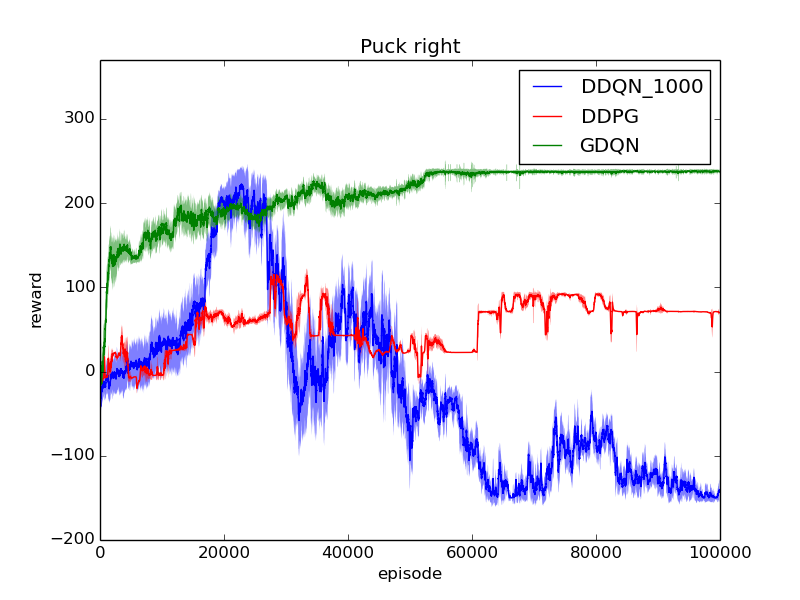}
}
\\
\framebox{
\includegraphics[scale=0.22]{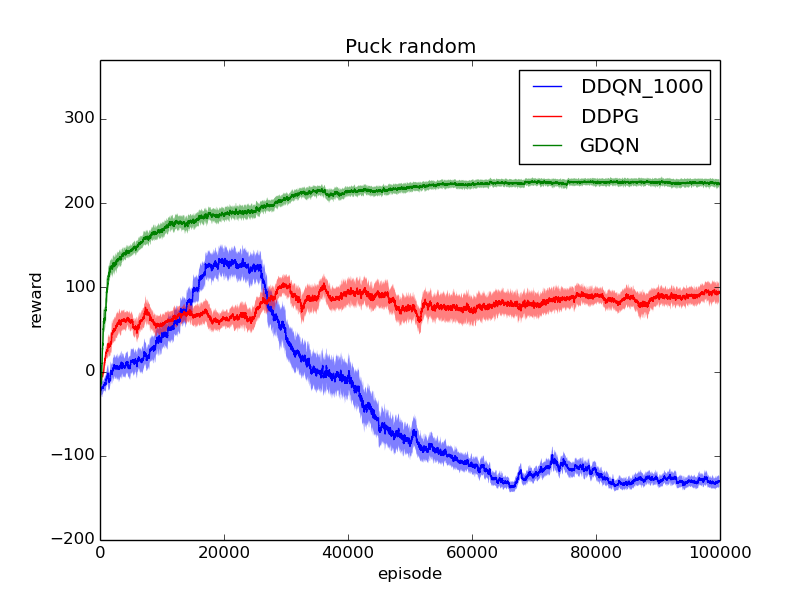}
}
\framebox{
\includegraphics[scale=0.22]{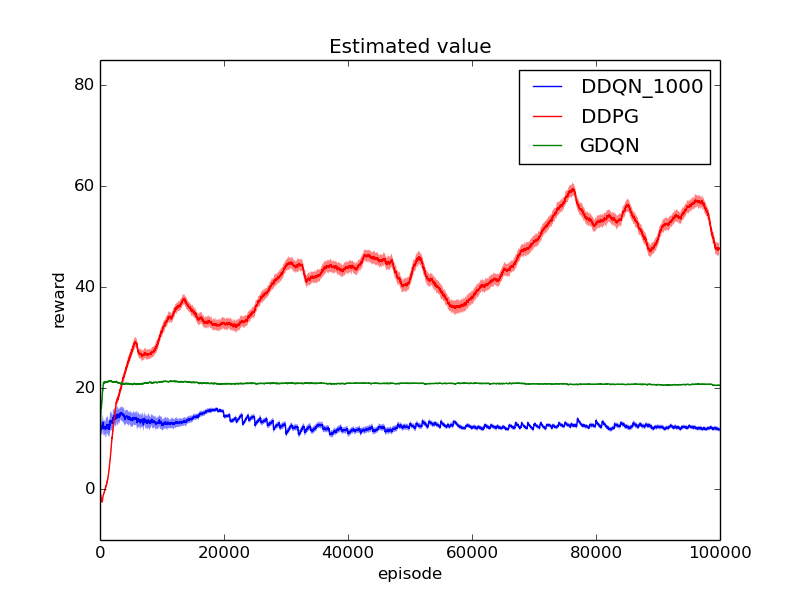}
}
      \caption{GDQN, DDQN, DDPG results for the air hockey striking problem.}
      \label{fig:All}
\end{figure*}

In Fig. \ref{fig:All} we compare the results of three algorithms, the DDQN algorithm with the intermediate update period (the best of the three shown before), Deep-mind's Deep Deterministic Policy Gradients (DDPG) algorithm, and our Guided-DQN algorithm. \\

The DDPG algorithm manages to learn a suboptimal policy, but oscillates strongly around it. It can be seen in the fix positions graphs of the puck, although in the random graphs it looks pretty stable on the suboptimal policy. DDQN was discussed before, and our GDQN as can be seen clearly, learns the optimal policy and reaches the maximum score possible for each of them. In the random puck position the score also reaches an optimal policy in a very stable manner. Note that the score doesn't drop at all, and even the rise at the beginning is faster than the other two algorithms, it even faster than the rise of the DDQN with the fast updates shown in Fig. \ref{fig:DDQN}, due to the fast updates at the beginning and the guidance of the teacher policy.
The average values of DDQN and DDPG are oscillating and suffering from underestimation and overestimation respectively, where GDQN's average value is extremely stable and does not suffer from over or under estimation. \\

\begin{figure}[h!]
\centering
\framebox{
\includegraphics[scale=0.36]{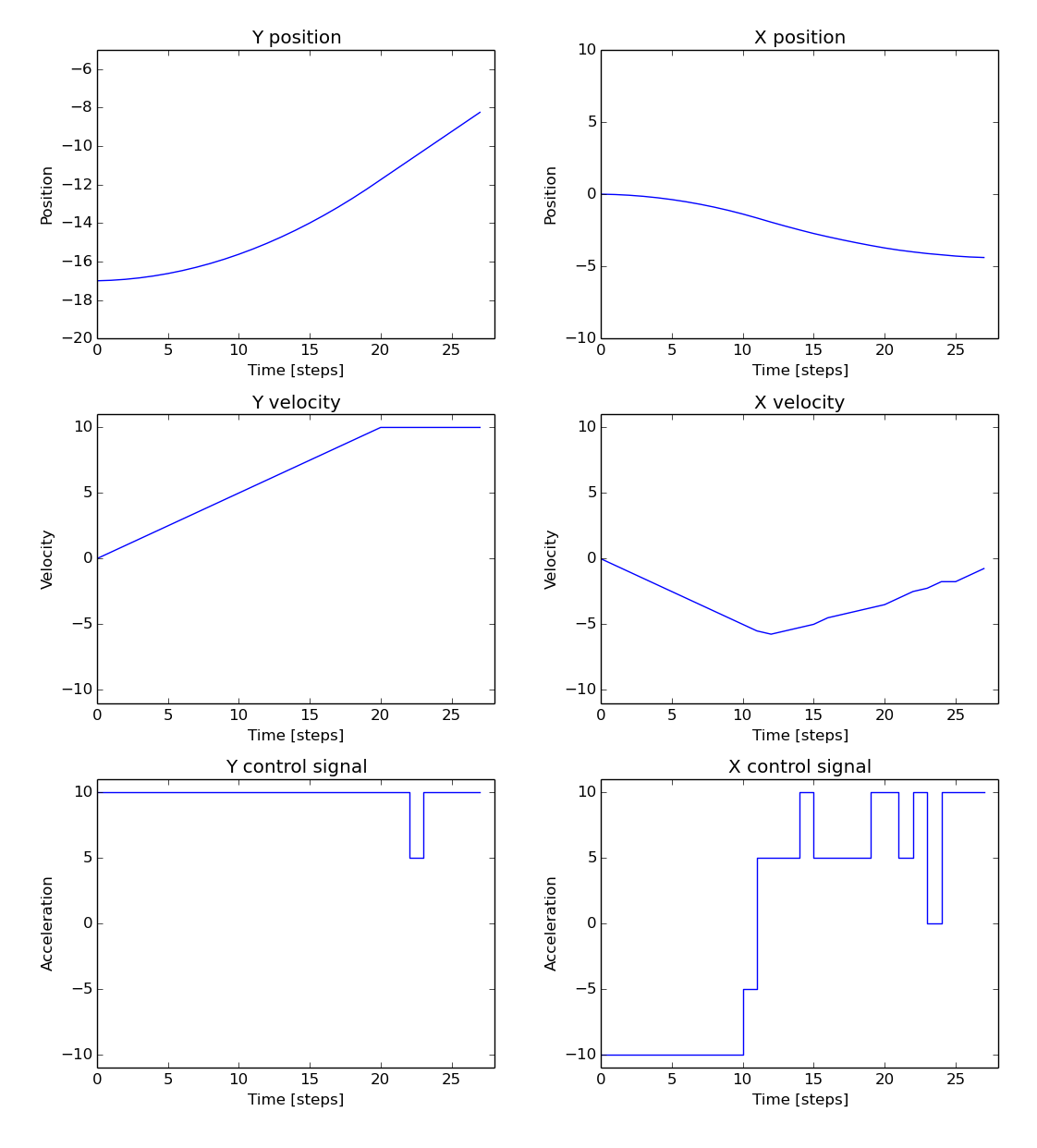}
}
\caption{Learned profiles for a puck stationed in the left side of the table. X axis is the horizontal dimension of the table, and Y axis the vertical dimension of the table. The first row is the position of the agent at each time step, the second row is the velocity profile of the agent and the third tow is the control signal sent to the agent.}
\label{fig:left_profile}
\end{figure}

\begin{figure}[!h]
\centering
\framebox{
\includegraphics[scale=0.52]{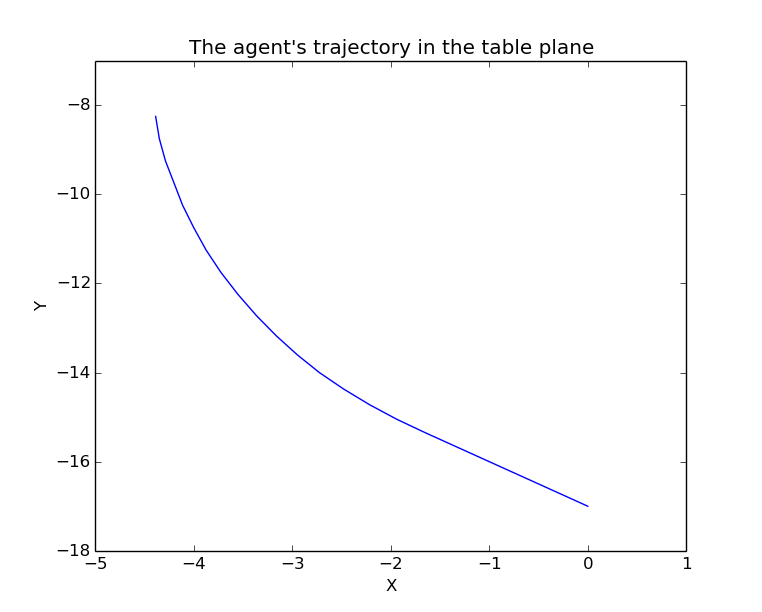}
}
\caption{The agent's trajectory in the X-Y plane of the table.}
\label{fig:left_trajectory}
\end{figure}

The learned control polices and the trajectories are shown in Fig. \ref{fig:left_profile} for a puck stationed in the left side of the table. The profile in the X-Y plane of the table is shown in Fig. \ref{fig:left_trajectory}. The agent is doing a curve in order to hit the puck from the left so it will go to the middle of the goal. The motion is visually very similar to an S-curve, in the X axis the agent performs a saturated action, compatible with a Bang-Bang profile, and in the Y axis something that effectively is like a Band-Zero-Bang.

\medskip

\section{\sc Conclusions}
\label{conclusions}
We addressed the application of striking a stationary puck with a physical mechanism. We showed that the standard DQN algorithm did not lead to satisfactory results. Therefore we proposed two novel improvements to this algorithm.
\begin{enumerate}
\item using prior knowledge during the learning to direct the algorithm to interesting region of the state and action spaces.
\item using non-uniform target update periods with expanding rate in order to stabilize the learning.
\end{enumerate}
We also augmented the commonly used $\epsilon$-greedy exploration mechanism with a local exploration with temporally correlated random process to better suite the physical environment. \\

The modified algorithm is shown to learn near optimal performance in the motion planning and control problem of air hockey striking.
In particular, it solves completely the problem of score drop that was observed in Double DQN.

\bibliographystyle{acm}
\bibliography{Bibliography}

\begin{thebibliography}{10}

\bibitem{abbeel2004apprenticeship}
{\sc Abbeel, P., and Ng, A.~Y.}
\newblock Apprenticeship learning via inverse reinforcement learning.
\newblock In {\em Proceedings of the twenty-first international conference on
  Machine learning\/} (2004), ACM, p.~1.

\bibitem{argall2009survey}
{\sc Argall, B.~D., Chernova, S., Veloso, M., and Browning, B.}
\newblock A survey of robot learning from demonstration.
\newblock {\em Robotics and autonomous systems 57}, 5 (2009), 469--483.

\bibitem{bentivegna2002framework}
{\sc Bentivegna, D.~C., and Atkeson, C.~G.}
\newblock A framework for learning from observation using primitives.
\newblock In {\em Robot Soccer World Cup\/} (2002), Springer, pp.~263--270.

\bibitem{brockman2016openai}
{\sc Brockman, G., Cheung, V., Pettersson, L., Schneider, J., Schulman, J.,
  Tang, J., and Zaremba, W.}
\newblock Openai gym.
\newblock {\em arXiv preprint arXiv:1606.01540\/} (2016).

\bibitem{duan2016benchmarking}
{\sc Duan, Y., Chen, X., Houthooft, R., Schulman, J., and Abbeel, P.}
\newblock Benchmarking deep reinforcement learning for continuous control.
\newblock {\em arXiv preprint arXiv:1604.06778\/} (2016).

\bibitem{french1999catastrophic}
{\sc French, R.~M.}
\newblock Catastrophic forgetting in connectionist networks.
\newblock {\em Trends in cognitive sciences 3}, 4 (1999), 128--135.

\bibitem{Kakade}
{\sc Kakade, S.~M.}
\newblock Truncated natural policy gradient.
\newblock {\em NIPS\/} (2002), 1531--1538.

\bibitem{kober2013reinforcement}
{\sc Kober, J., Bagnell, J.~A., and Peters, J.}
\newblock Reinforcement learning in robotics: A survey.
\newblock {\em The International Journal of Robotics Research\/} (2013),
  0278364913495721.

\bibitem{lillicrap2015continuous}
{\sc Lillicrap, T.~P., Hunt, J.~J., Pritzel, A., Heess, N., Erez, T., Tassa,
  Y., Silver, D., and Wierstra, D.}
\newblock Continuous control with deep reinforcement learning.
\newblock {\em arXiv preprint arXiv:1509.02971\/} (2015).

\bibitem{mnih2016asynchronous}
{\sc Mnih, V., Badia, A.~P., Mirza, M., Graves, A., Lillicrap, T.~P., Harley,
  T., Silver, D., and Kavukcuoglu, K.}
\newblock Asynchronous methods for deep reinforcement learning.
\newblock {\em arXiv preprint arXiv:1602.01783\/} (2016).

\bibitem{mnih2013playing}
{\sc Mnih, V., Kavukcuoglu, K., Silver, D., Graves, A., Antonoglou, I.,
  Wierstra, D., and Riedmiller, M.}
\newblock Playing atari with deep reinforcement learning.
\newblock {\em arXiv preprint arXiv:1312.5602\/} (2013).

\bibitem{mnih2015human}
{\sc Mnih, V., Kavukcuoglu, K., Silver, D., Rusu, A.~A., Veness, J., Bellemare,
  M.~G., Graves, A., Riedmiller, M., Fidjeland, A.~K., Ostrovski, G., et~al.}
\newblock Human-level control through deep reinforcement learning.
\newblock {\em Nature 518}, 7540 (2015), 529--533.

\bibitem{muelling2010learning}
{\sc Muelling, K., Kober, J., and Peters, J.}
\newblock Learning table tennis with a mixture of motor primitives.
\newblock In {\em 2010 10th IEEE-RAS International Conference on Humanoid
  Robots\/} (2010), IEEE, pp.~411--416.

\bibitem{nair2015massively}
{\sc Nair, A., Srinivasan, P., Blackwell, S., Alcicek, C., Fearon, R.,
  De~Maria, A., Panneershelvam, V., Suleyman, M., Beattie, C., Petersen, S.,
  et~al.}
\newblock Massively parallel methods for deep reinforcement learning.
\newblock {\em arXiv preprint arXiv:1507.04296\/} (2015).

\bibitem{namiki2013}
{\sc Namiki, A., Matsushita, S., Ozeki, T., and Nonami, K.}
\newblock Hierarchical processing architecture for an air-hockey robot system.
\newblock In {\em Robotics and Automation (ICRA), 2013 IEEE International
  Conference on\/} (2013), IEEE, pp.~1187--1192.

\bibitem{partridge2000control}
{\sc Partridge, C.~B., and Spong, M.~W.}
\newblock Control of planar rigid body sliding with impacts and friction.
\newblock {\em The International Journal of Robotics Research 19}, 4 (2000),
  336--348.

\bibitem{peters2007reinforcement}
{\sc Peters, J., and Schaal, S.}
\newblock Reinforcement learning by reward-weighted regression for operational
  space control.
\newblock In {\em Proceedings of the 24th international conference on Machine
  learning\/} (2007), ACM, pp.~745--750.

\bibitem{schaal1997learning}
{\sc Schaal, S.}
\newblock Learning from demonstration.
\newblock {\em Advances in neural information processing systems\/} (1997),
  1040--1046.

\bibitem{schaul2015prioritized}
{\sc Schaul, T., Quan, J., Antonoglou, I., and Silver, D.}
\newblock Prioritized experience replay.
\newblock {\em arXiv preprint arXiv:1511.05952\/} (2015).

\bibitem{schulman2015trust}
{\sc Schulman, J., Levine, S., Moritz, P., Jordan, M.~I., and Abbeel, P.}
\newblock Trust region policy optimization.
\newblock {\em CoRR, abs/1502.05477\/} (2015).

\bibitem{schulman2015high}
{\sc Schulman, J., Moritz, P., Levine, S., Jordan, M., and Abbeel, P.}
\newblock High-dimensional continuous control using generalized advantage
  estimation.
\newblock {\em arXiv preprint arXiv:1506.02438\/} (2015).

\bibitem{silver2016mastering}
{\sc Silver, D., Huang, A., Maddison, C.~J., Guez, A., Sifre, L., Van
  Den~Driessche, G., Schrittwieser, J., Antonoglou, I., Panneershelvam, V.,
  Lanctot, M., et~al.}
\newblock Mastering the game of go with deep neural networks and tree search.
\newblock {\em Nature 529}, 7587 (2016), 484--489.

\bibitem{uhlenbeck1930theory}
{\sc Uhlenbeck, G.~E., and Ornstein, L.~S.}
\newblock On the theory of the brownian motion.
\newblock {\em Physical review 36}, 5 (1930), 823.

\bibitem{van2015deep}
{\sc Van~Hasselt, H., Guez, A., and Silver, D.}
\newblock Deep reinforcement learning with double q-learning.
\newblock {\em CoRR, abs/1509.06461\/} (2015).

\bibitem{watkins1992q}
{\sc Watkins, C.~J., and Dayan, P.}
\newblock Q-learning.
\newblock {\em Machine learning 8}, 3-4 (1992), 279--292.

\bibitem{williams1992simple}
{\sc Williams, R.~J.}
\newblock Simple statistical gradient-following algorithms for connectionist
  reinforcement learning.
\newblock {\em Machine learning 8}, 3-4 (1992), 229--256.

\end{thebibliography}

\end{document}